\documentclass{article}

\PassOptionsToPackage{numbers, compress}{natbib}

\usepackage[final]{neurips_2019}

\usepackage[utf8]{inputenc} 
\usepackage[T1]{fontenc}    
\usepackage{hyperref}       
\usepackage{url}            
\usepackage{booktabs}       
\usepackage{amsfonts}       
\usepackage{nicefrac}       
\usepackage{microtype}      
\usepackage{amsmath,amsthm,amssymb,bm,mathtools}
\usepackage{graphicx}
\graphicspath{{fig/}}

\newcommand{\tr}{^\intercal}
\newcommand{\email}[1]{\href{mailto:#1}{\nolinkurl{#1}}}

\DeclareMathOperator*{\argmax}{arg\,max}

\newcommand\COtwo{CO\textsubscript{2}}
\newcommand\given{\;|\;}
\renewcommand{\vec}[1]{\boldsymbol{#1}}
\newenvironment{enumerateb}
  {\begin{enumerate}}
 {\end{enumerate}}
\newcommand{\actiontitle}[1]{\textbf{#1}}
\newcommand{\actionvalue}[1]{\textbf{Carbon footprint:} #1 kg\COtwo-equivalent.}

\title{A User Study of Perceived Carbon Footprint}

\author{%
  Victor Kristof\\
  EPFL
  \And
  Valentin Quelquejay-Leclère \\
  EPFL
  \And
  Robin Zbinden \\
  EPFL
  \AND
  Lucas Maystre \\
  Spotify
  \And
  Matthias Grossglauser \\
  EPFL
  \And
  Patrick Thiran \\
  EPFL
}

\begin{document}

\maketitle

\begin{abstract}
  We propose a statistical model to understand people's perception of their carbon footprint.
  Driven by the observation that few people think of \COtwo\ impact in absolute terms, we design a system to probe people's perception from simple pairwise comparisons of the relative carbon footprint of their actions.
  The formulation of the model enables us to take an active-learning approach to selecting the pairs of actions that are maximally informative about the model parameters.
  We define a set of 18 actions and collect a dataset of 2183 comparisons from 176 users on a university campus.
  The early results reveal promising directions to improve climate communication and enhance climate mitigation.
\end{abstract}

\section{Introduction}
\label{sec:introduction}
To put the focus on actions that have high potential for emission reduction, we must first understand whether people have an accurate perception of the carbon footprint of these actions.
If they do not, their efforts might be wasted.
As an example, recent work by \citet{wynes2017climate} shows that Canadian high-school textbooks encourage daily actions that yield negligible emission reduction.
Actions with a higher potential of emission reduction are poorly documented.
In this work, we model how people perceive the carbon footprint of their actions, which could guide educators and policy-makers.

In their daily life, consumers repeatedly face multiple options with varying environmental effects.
Except for a handful of experts, no one is able to estimate the absolute quantity of \COtwo\ emitted by their actions of say, flying from Paris to London.
Most people, however, are aware that taking the train for the same trip would release less \COtwo.
Hence, in the spirit of \citet{thurstone1927method} and \citet{salganik2015wiki} (among many others), we posit that the perception of a population can be probed by simple pairwise comparisons.
By doing so, we shift the complexity from the probing system to the model: Instead of asking difficult questions about each action and simply averaging the answers, we ask simple questions in the form of comparisons and design a non-trivial model to estimate the perception.
\textit{In fine}, human behaviour boils down to making choices: For example, we choose between eating local food and eating imported food; we do not choose between eating or not eating.
Our awareness of relative emissions between actions (of the same purpose) is often sufficient to improve our carbon footprint.

Our contributions are as follows.
First, we cast the problem of inferring a population's global perception from pairwise comparisons as a linear regression.
Second, we adapt a well-known active-learning method to maximize the information gained from each comparison.
We describe the model and the active-learning algorithm in Section~\ref{sec:model}.
We design an interactive platform to collect real data for an experiment on our university campus, and we show early results in Section~\ref{sec:results}.
Our approach could help climate scientists, sociologists, journalists, governments, and individuals improve climate communication and enhance climate mitigation.

\section{Model}%
\label{sec:model}

Let $ \mathcal{A} $ be a set of $M$ actions.
For instance, "flying from London to New York" or "eating meat for a year" are both actions in $ \mathcal{A}$.
Let $ (i, j, y) $ be a triplet encoding that action $i \in \mathcal{A}$ has an \textit{impact ratio} of $y \in \mathbf{R}_{>0}$ over action $j \in \mathcal{A}$.
Said otherwise, if $y > 1$, action $i$ has a carbon footprint $y$ times \textit{greater} than action $j$, and if $y < 1$, action $i$ has a carbon footprint $1/y$ times \textit{smaller} than action $j$.

Given some parameters $w_i, w_j \in \mathbf{R}$ representing the perceived (log-)carbon footprint in \COtwo-equivalent of action $i$ and action $j$, we posit
\begin{equation*}
   y = \frac{\exp w_i}{\exp w_j}.
\end{equation*}
We gather the parameters in a vector $\vec{w} \in \mathbf{R}^M$.
Assuming a centered Gaussian noise $\epsilon \sim \mathcal{N}(0, \sigma^2_n)$, $\sigma_n^2 \in \mathbf{R}$, we model the (log-)impact~ratio
\begin{equation}
  \log y = w_i - w_j + \epsilon = \vec{x}\tr\vec{w} + \epsilon,
\end{equation}
where the comparison vector $\vec{x} \in \mathbf{R}^M$ is zero everywhere except in entry~$i$ where it is $+1$ and in entry~$j$ where it is $-1$.
Vector $\vec{x}$ "selects" the pair of actions to compare.
For a dataset $ \mathcal{D} = \{ (i_n, j_n, y_n) : n = 1, ..., N \}$ of $N$ independent triplets and since \mbox{$\log y \sim \mathcal{N}(\vec{x}\tr \vec{w}, \sigma^2_n)$}, the likelihood of the model is
\begin{equation*}
  p(\vec{y} \given \vec{X}, \vec{w}) = \prod_{i=1}^N p(y_i \given \vec{x}_i\tr \vec{w}, \sigma_n^2) = \mathcal{N}(\vec{X} \vec{w}, \sigma_n^2\vec{I}),
\end{equation*}
where $\vec{y} \in \mathbf{R}^N$ is the vector of observed (log-)impact ratios, and $\vec{X} \in \mathbf{R}^{N \times M}$ is a matrix of $N$ comparison vectors.

We assume a Gaussian prior for the weight parameters $\vec{w} \sim \mathcal{N}(\vec{\mu}, \vec{\Sigma}_p)$, where $\vec{\mu} \in \mathbf{R}^M $ is the prior mean and $\vec{\Sigma}_p \in \mathbf{R}^{M \times M}$ is the prior covariance matrix.
To obtain the global perceived carbon footprint of each action in $\mathcal{A}$ and to enable active learning, we compute the posterior distribution of the weight parameters given the data,
\begin{eqnarray}
  \label{eq:posterior}
  p(\vec{w} \given \vec{X}, \vec{y})
  &=& \frac{p(\vec{y} \given \vec{X}, \vec{w})p(\vec{w})}{p(\vec{y} \given \vec{X})} \nonumber \\
  &=& \mathcal{N}\left(\overline{\vec{w}} = \vec{\Sigma} \left( \sigma_n^{-2} \vec{X}\tr \vec{y} + \vec{\Sigma}_p^{-1} \vec{\mu}\right), \vec{\Sigma} = \left( \sigma_n^{-2} \vec{X}\tr \vec{X} + \vec{\Sigma}_p^{-1} \right)^{-1}\right).
\end{eqnarray}
The noise variance $\sigma_n^2$, the prior mean $\vec{\mu}$, and the prior covariance matrix $\vec{\Sigma}_p$ are hyperparameters to be tuned.
The global perceived carbon footprint is given by the posterior mean as $\exp \overline{\vec{w}}$.
We use the posterior covariance matrix $\vec{\Sigma}$ to select the next pair of actions, as described in the following section.

\paragraph{Active Learning}
We collect the triplets in $\mathcal{D}$ from multiple users who take a quiz.
During one session of the quiz, a user sequentially answers comparison questions and decides when to stop to see their overall results.
Active learning enables us to maximize the information extracted from a session.

Let $\vec{\Sigma}_N$ and  $\vec{\Sigma}_{N+1}$ be the covariance matrices of the posterior distribution in Equation~\eqref{eq:posterior} when $N$ and $N+1$ comparisons have been respectively collected.
Let $\vec{x}$ be the new $(N+1)$-th comparison vector.
As proposed by~\citet{mackay1992information}, we want to select the pair of actions to compare that is maximally informative about the values that the model parameters $\vec{w}$ should take \citep{chu2005extensions, houlsby2012collaborative}.
For our linear Gaussian model, this is obtained by maximizing the total information gain
\begin{equation}
  \Delta S = S_N - S_{N+1} = \frac{1}{2} \log (1 + \sigma_n^{-2} \vec{x}\tr \vec{\Sigma}_N \vec{x}),
\end{equation}
where $S_N = \frac{M}{2}(1 + \log 2 \pi) + \frac{1}{2} \log \det \vec{\Sigma}_N$ is the entropy of a multivariate Gaussian distribution with covariance matrix $\vec{\Sigma}_N$.
The full derivation of the total information gain can be found in Appendix~\ref{app:active_learning}.
To maximize $\Delta S$, we maximize $ \vec{x}\tr \vec{\Sigma}_N \vec{x} $ for all possible $ \vec{x} $ in our dataset.
Recall that comparison vectors $ \vec{x} $ are zero everywhere except in entry~$i$~(+1) and in entry~$j$~(-1).
By denoting $\vec{\Sigma}_N = [\sigma_{ij}^2]_{i, j = 1}^M$, we seek, therefore, to find the pair of actions
\begin{equation*}
  (i^\star, j^\star) = \argmax_{i, j} \left\{ \sigma_{ii}^2 + \sigma_{jj}^2 - 2 \sigma_{ij}^2 \right\}.
\end{equation*}

The prior covariance matrix $\vec{\Sigma}_p$ could capture the prior knowledge about the typical user perception of relative carbon footprint.
In future work, we intend to further reduce the number of questions asked during one session by a judicious choice of $\vec{\Sigma}_p$.
In our experiments so far, we simply initialize it to a spherical covariance, as explained in the next section.

\section{Results}
\label{sec:results}

Starting with no information at all, we arbitrarily set the prior noise $\sigma_n^{2} = 1$ and the prior covariance matrix to a spherical covariance $\vec{\Sigma}_p = \sigma_p^{2} \vec{I}$, with $\sigma_p^{2} = 10$.
Our results are qualitatively robust to a large range of values for $\sigma_p^{2}$.
In order to compare the perceived carbon footprint $\exp \overline{\vec{w}}$ with its true value $\exp \vec{v}$, we set the prior mean to $\vec{\mu} = c \vec{1}$, where $ c = \frac{1}{M}\sum_{i=1}^M v_i$ is the mean of the (log-)true values.
This guarantees that the perceived carbon footprint estimated from the model parameters have the same scale as the true values.

\begin{figure}
  \centering
  \includegraphics{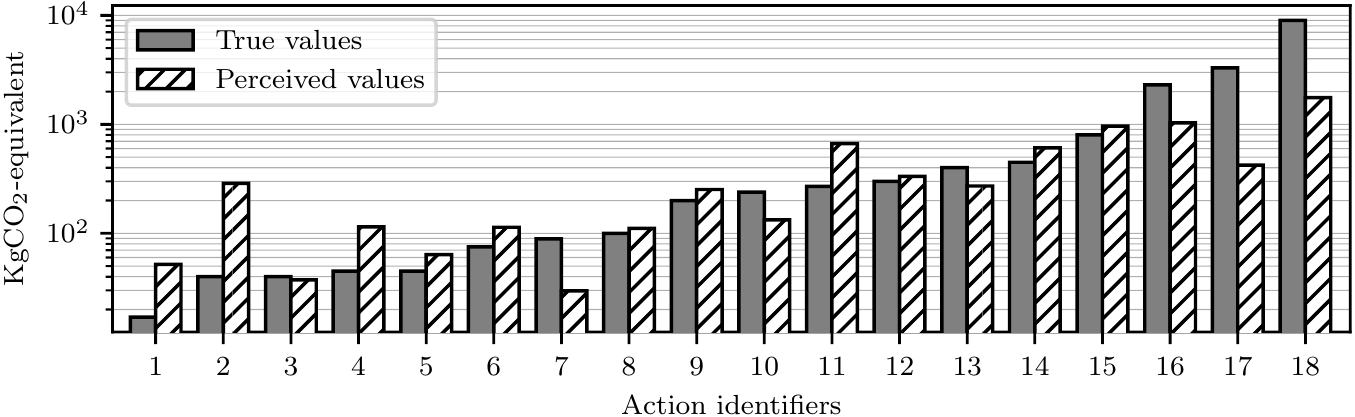}
  \caption{Global perceived carbon footprint of 18 actions in kg\COtwo-equivalent and their true values (log scale).
  The list of actions is provided in Appendix~\ref{app:actions}.}%
  \label{fig:perception}
\end{figure}

We compile a set $\mathcal{A}$ of $M=18$ individual actions about transportation, food, and household (the full list of actions is provided in Appendix~\ref{app:actions}).
We deploy an online quiz\footnote{Accessible at \url{http://www.climpact.ch}} to collect pairwise comparisons of actions from real users on a university campus.
We collect $N=2183$ triplets from 176 users, mostly students between 16 and 25 years old.
We show in Figure~\ref{fig:perception} the true carbon footprint, together with the global perception of the population, i.e., the values $\exp \overline{w}_i$ for each action $i \in \mathcal{A}$.

The users in our population have a globally accurate perception.
Among the actions showing the most discrepancy, the carbon footprint of short-haul flights is \textit{overestimated}~(Action 11), whereas the carbon footprint of long-haul flights~(16) is highly \textit{underestimated}~(the scale is logarithmic).
Similarly, the carbon footprint of first-class flights~(18) is also \textit{underestimated}.
The users tend to \textit{overestimate} the carbon footprint of more ecological transports, such as the train, the bus, and car-sharing~(1, 4, and 6).
The users have an accurate perception of actions related to diet~(8, 14, and 15) and of actions related to domestic lighting~(3 and 10).
They \textit{overestimate}, however, the carbon footprint of a dryer~(2).
Finally, they highly \textit{underestimate} the carbon footprint of oil heating~(17).
Switzerland, where the users live, is one of the European countries whose consumption of oil for heating houses is the highest.
There is, therefore, a high potential for raising awareness around this issue.

\section{Conclusion}
\label{sec:conclusion}

In this work, we proposed a statistical model for understanding people's global perception of their carbon footprint.
The Bayesian formulation of the model enables us to take an active-learning approach to selecting the pairs of actions that maximize the gain of information.
We deployed an online platform to collect real data from users.
The estimated perception of the users gives us insight into this population and reveals interesting directions for improving climate communication.

For future work, we will open and deploy our platform to a wider audience.
We plan to collaborate with domain experts to further analyze people's estimated perception of their carbon footprint and to translate the conclusions of the results into concrete actions.

\newpage
\bibliographystyle{abbrvnat}
\bibliography{climpact}

\appendix
\section{Appendix}



\subsection{Total Information Gain for Multivariate Gaussian Distributions}
\label{app:active_learning}

Recall that the entropy of a multivariate Gaussian distribution $ \mathcal{N}(\vec{\mu}, \vec{\Sigma}) $, $\vec{\mu} \in \mathbf{R}^M$, $\vec{\Sigma} \in \mathbf{R}^{M \times M}$, is given by
\begin{equation}
  S = \frac{M}{2}(1 + \log 2 \pi) + \frac{1}{2} \log \det \vec{\Sigma}.
\end{equation}
Let $\vec{\Sigma}_N$ and  $\vec{\Sigma}_{N+1}$ be the covariance matrices of the posterior distribution in Equation~\eqref{eq:posterior} when $N$ and $N+1$ data points have been collected, respectively.
Let $\vec{x}$ be the new $(N+1)$-th data point.
The total information gain is
\begin{eqnarray}
  \Delta S &=& S_N - S_{N+1}  \nonumber \\
           &=& \frac{1}{2} \log \frac{\det \vec{\Sigma}_{N+1}^{-1}}{\det \vec{\Sigma}_N^{-1}} \nonumber \\
           &=& \frac{1}{2} \log \frac{\det [\vec{\Sigma}_N^{-1} + \sigma_n^{-2}\vec{x}\vec{x}\tr] }{\det \vec{\Sigma}_N^{-1}} \label{eq:covariance} \\
           &=& \frac{1}{2} \log \frac{(\det \vec{\Sigma}_N^{-1})(1 + \sigma_n^{-2} \vec{x}\tr \vec{\Sigma}_N \vec{x} ) }{ \det \vec{\Sigma}_N^{-1} } \label{eq:determinant} \\
           &=& \frac{1}{2} \log (1 + \sigma_n^{-2} \vec{x}\tr \vec{\Sigma}_N \vec{x}). \nonumber
\end{eqnarray}
We obtain Equation~\eqref{eq:covariance} by observing that $ \vec{\Sigma}_N^{-1} + \sigma_n^{-2} \vec{x} \vec{x}\tr = \sigma_n^{-2} \vec{X}\tr \vec{X} + \vec{\Sigma}_p^{-1} + \sigma_n^{-2} \vec{x} \vec{x}\tr = \vec{\Sigma}_{N+1}^{-1}$.
We obtain Equation~\eqref{eq:determinant} by the matrix determinant lemma.

\subsection{List of Actions}
\label{app:actions}

We provide here the full list of actions, together with the true carbon footprint associated with each of them.
Because different countries use different sources of energy, we calculate the carbon footprint \textit{relative} to the country where our university is located.
The actions are ordered according to their true carbon footprint.
\begin{enumerateb}
  \item \actiontitle{Take the train in economy class on a 1000-km round-trip.} \\
        The train is a high-speed train with 360 seats.
        The seat-occupancy rate is 55\% (average rate for these types of trains).
        We count the \COtwo\ emissions per passenger. \\
        \actionvalue{17} 
  \item \actiontitle{Dry your clothes with a dryer for one year.} \\
        A dryer emits \COtwo\ because it consumes electricity.
        We consider a dryer of average quality.
        The electricity is consumed from a grid with average \COtwo\ rate. \\
        \actionvalue{40} 
  \item \actiontitle{Light your house with LED bulbs.} \\
        LED bulbs emit CO2 because they consume electricity to generate light.
        The electricity is consumed from a grid with average \COtwo\ rate. \\
        \actionvalue{40} 
  \item \actiontitle{Take the bus on a 1000-km round-trip.} \\
        The bus is a standard-size bus with 60 seats.
        The seat-occupancy rate is 50\% (average rate for buses).
        We count the \COtwo\ emissions per passenger. \\
        \actionvalue{45} 
  \item \actiontitle{Drive an electric car alone on a 1000-km round-trip.} \\
        The car is a compact electric car that consumes 15 kWh/100km.
        The electricity is consumed from a grid with average \COtwo\ rate.
        There are no other passengers in the car.
        We count the \COtwo\ emissions per passenger. \\
        \actionvalue{45} 
  \item \actiontitle{Car-share with three other persons on a 1000-km round-trip.} \\
        The car is a mid-sized gasoline car that consumes 7 l/100km.
        There are four persons in the car.
        We count the \COtwo\ emissions per passenger. \\
        \actionvalue{75} 
  \item \actiontitle{Eat local and seasonal fruits and vegetables for one year.} \\
        Growing food emits \COtwo\ because it requires fertilizing and driving agricultural machines.
        The goods are then transported to grocery shops and to your home. \\
        \actionvalue{89} 
  \item \actiontitle{Eat eggs and dairy products for one year.} \\
        The production of eggs and dairy products (milk, cheese, etc.) emits \COtwo\ because of water and land consumption, animal methane, and fossil fuel consumption for transportation and heating.
        We consider an average citizen consuming 50 kg of eggs and dairy products per year. \\
        \actionvalue{100} 
  \item \actiontitle{Throw all waste in the same trash for one year.} \\
        Throwing all waste (PET, glass, cardboard, etc.) in the same trash, i.e., without recycling, emits \COtwo\ because more energy is needed to extract, transport, and process raw materials.
        Incinerators also burn more waste, and organic waste decomposition generates methane. \\
        \actionvalue{200} 
  \item \actiontitle{Light your house with incandescent bulbs.} \\
        Incandescent bulbs emit \COtwo\ because they consume electricity to generate light.
        The electricity is consumed from a grid with average \COtwo\ rate. \\
        \actionvalue{239} 
  \item \actiontitle{Fly in economy class for a 800-km round-trip.} \\
        The plane is a standard aircraft for short-distance flights with 180 seats.
        The seat-occupancy rate is 80\%.
        We count the \COtwo\ emissions per passenger. \\
        \actionvalue{270} 
  \item \actiontitle{Drive alone for a 1000-km round-trip.} \\
        The car is a mid-sized gasoline car that consumes 7 l/100km.
        There are no other passengers in the car.
        We count the \COtwo\ emissions per passenger. \\
        \actionvalue{300} 
  \item \actiontitle{Heat your house with a heat pump for one year.} \\
        A heat pump emits \COtwo\ because it consumes electricity to generate heat.
        The house is of average size.
        The electricity is consumed from a grid with average \COtwo\ rate. \\
        \actionvalue{400} 
  \item \actiontitle{Eat imported and out-of-season fruits and vegetables for one year.} \\
        Growing food emits \COtwo\  because it requires fertilizing and driving agricultural machines.
        Importing food emits \COtwo\ because of fossil fuel consumption for transportation.
        Out-of-season food emits \COtwo\ because it grows in greenhouse that needs to be heated.
        The goods are then transported to grocery shops and to your home. \\
        \actionvalue{449} 
  \item \actiontitle{Eat meat for one year.} \\
        Meat production emits \COtwo\ because of water and land consumption, animal methane, and fossil fuel consumption for transportation and heating.
        We consider an average citizen consuming 50 kg of meat per year. \\
        \actionvalue{800} 
  \item \actiontitle{Fly in economy class for a 12000-km round-trip.} \\
        The plane is a standard aircraft for long-distance flights with 390 seats.
        The seat-occupancy rate is close to 100\%.
        We count the \COtwo\ emissions per passenger. \\
        \actionvalue{2300} 
  \item \actiontitle{Heat your house with an oil furnace for one year.} \\
        An oil furnace emits \COtwo\ because it burns fuel to generate heat.
        The house is of average size. \\
        \actionvalue{3300} 
  \item \actiontitle{Fly in first class for a 12000-km round-trip.} \\
        The plane is a standard aircraft for long-distance flights with 390 seats.
        The seat-occupancy rate is close to 100\%.
        We count the \COtwo\ emissions per passenger.
        Passengers flying in first class use more space than passengers in economy. \\
        \actionvalue{9000} 
\end{enumerateb}

\end{document}